%% file: root.tex
\documentclass[letterpaper, 10 pt, conference]{ieeeconf}  % Comment this line out if you need a4paper

\IEEEoverridecommandlockouts                              % This command is only needed if 
\usepackage{amsmath} % assumes amsmath package installed

\usepackage{amssymb}  % assumes amsmath package installed
\usepackage{bbm}
\usepackage[bb=boondox]{mathalfa}
\usepackage{graphicx} % for pdf, bitmapped graphics files
\usepackage{hyperref}
\usepackage{booktabs,arydshln}
\makeatletter
\def\adl@drawiv#1#2#3{%
        \hskip.5\tabcolsep
        \xleaders#3{#2.5\@tempdimb #1{1}#2.5\@tempdimb}%
                #2\z@ plus1fil minus1fil\relax
        \hskip.5\tabcolsep}
\newcommand{\cdashlinelr}[1]{%
  \noalign{\vskip\aboverulesep
           \global\let\@dashdrawstore\adl@draw
           \global\let\adl@draw\adl@drawiv}
  \cdashline{#1}
  \noalign{\global\let\adl@draw\@dashdrawstore
           \vskip\belowrulesep}}
\makeatother

\usepackage{threeparttable}
\usepackage{tabularx}
\usepackage{multirow}
\usepackage[caption=false, font=footnotesize]{subfig}
\usepackage{algorithmic}
\usepackage{algorithm}

\usepackage{color}
\usepackage{xcolor}
\usepackage{colortbl}
\definecolor{green}{rgb}{0.8,1.0,0.8}
\definecolor{red}{rgb}{1.0,0.8,0.8}

\usepackage{comment}
\usepackage[%
backend=biber,
url=true,
block=space,
maxbibnames=10,
minbibnames=1,
bibstyle=ieee,
]{biblatex}
\AtBeginBibliography{\footnotesize}
\AtEveryBibitem{
  \clearfield{doi}
  \clearfield{isbn}
  \clearfield{issn}
  \clearfield{month}
  \clearfield{publisher}
  \ifentrytype{inproceedings}{
    \clearfield{arxivId}
    \clearfield{archivePrefix}
    \clearfield{eprint}
    \clearfield{url}
  }{}
  \ifentrytype{article}{
    \clearfield{arxivId}
    \clearfield{archivePrefix}
    \clearfield{eprint}
    \clearfield{url}
  }{}
  \ifentrytype{report}{
  }{}
}
\renewbibmacro{in:}{%
  \ifentrytype{inproceedings}{%
    \setunit{}
    \addperiod\addspace In \textit{Proc.\ of the}}%
  {\printtext{\bibstring{in}\intitlepunct}}
}
\DeclareSourcemap{
  \maps{
    \map{ % Replaces '{\_}', '{_}' or '\_' with just '_'
      \step[fieldsource=url,
      match=\regexp{\{\\\_\}|\{\_\}|\\\_},
      replace=\regexp{\_}]
    }
    \map{ % Replaces '{'$\sim$'}', '$\sim$' or '{~}' with just '~'
      \step[fieldsource=url,
      match=\regexp{\{\$\\sim\$\}|\{\~\}|\$\\sim\$},
      replace=\regexp{\~}]
    }
    \map{ % Replaces '{\$}'
      \step[fieldsource=url,
      match=\regexp{\{\\\x{26}\}},
      replace=\regexp{\x{26}}]
    }
  }
}
\title{\LARGE \bf
CLIP-Loc: Multi-modal Landmark Association for \\ Global Localization in Object-based Maps
}

\author{Shigemichi Matsuzaki, Takuma Sugino, Kazuhito Tanaka, Zijun Sha, \\
Shintaro Nakaoka, Shintaro Yoshizawa, and Kazuhiro Shintani% <-this % stops a space
\thanks{Authors are with Frontier Research Center, Toyota Motor Corporation,
Toyota, Aichi, Japan.
}%
\thanks{{\tt\small shigemichi\_matsuzaki@mail.toyota.co.jp}}
}

\begin{document}

\maketitle
\thispagestyle{empty}
\pagestyle{empty}
\begin{comment}
物体ベース地図とカメラ画像を用いた大域的位置推定のためのマルチモーダルデータ対応付け手法を提案する．
物体ベース地図を用いた大域的位置推定やリローカライゼーションでは，
同一カテゴリの検出物体とランドマークのすべての可能な組み合わせ
を対応候補とし，RANSACや総当たり法等でインライヤを求める手法が主として用いられる．
この手法はランドマーク数が増加するにつれて対応候補数が爆発的に増加するためスケーラブルでない．
本論文では，ランドマークを自然言語による記述でラベル付けし，Vision Language Model (VLM)
を用いて画像観測との類似度に基づいて対応を抽出する手法を提案する．
言語による詳細な情報を用いることで，カテゴリ情報のみを用いる手法より効率的に対応を抽出することが可能になる．
デスク一つの規模から一部屋規模の異なるデータセットを用いた実験により，
提案手法が物体カテゴリに基づくベースライン手法と比較し
より少ない反復回数で正確な位置推定が可能であることを示す．
\end{comment}
%%%%%%%%%%%%%%%%%%%%%%%%%%%%%%%%%%%%%%%%%%%%%%%%%%%%%%%%%%%%%%%%%%%%%%%%%%%%%%%%
\begin{abstract}
	This paper describes a multi-modal data association method for global localization
	using object-based maps and camera images.
	In global localization, or relocalization, using object-based maps,
	existing methods typically resort to matching all possible combinations of
	detected objects and landmarks with the same object category,
	followed by inlier extraction using RANSAC or brute-force search.
	This approach becomes infeasible as the number of landmarks
	increases due to the exponential growth of correspondence candidates.
	In this paper, we propose labeling landmarks with natural language descriptions and
	extracting correspondences based on conceptual similarity with image observations
	using a Vision Language Model (VLM).
	By leveraging detailed text information,
	our approach efficiently extracts correspondences
	compared to methods using only object categories.
	Through experiments, %conducted on datasets ranging from desk-scale to single-room scale,
	we demonstrate that the proposed method enables more accurate global localization
	with fewer iterations compared to baseline methods, exhibiting its efficiency.
\end{abstract}

% This paper describes a multi-modal data association method for global localization using object-based maps and camera images. In global localization, or relocalization, using object-based maps, existing methods typically resort to matching all possible combinations of detected objects and landmarks with the same object category, followed by inlier extraction using RANSAC or brute-force search. This approach becomes infeasible as the number of landmarks increases due to the exponential growth of correspondence candidates. In this paper, we propose labeling landmarks with natural language descriptions and extracting correspondences based on conceptual similarity with image observations using a Vision Language Model (VLM). By leveraging detailed text information, our approach efficiently extracts correspondences compared to methods using only object categories. Through experiments, we demonstrate that the proposed method enables more accurate global localization with fewer iterations compared to baseline methods, exhibiting its efficiency.

%%%%%%%%%%%%%%%%%%%%%%%%%%%%%%%%%%%%%%%%%%%%%%%%%%%%%%%%%%%%%%%%%%%%%%%%%%%%%%%%
\input{section/1_introduction.tex}

\input{section/2_related_work.tex}
\input{section/3_proposed_method.tex}

\input{section/4_experiment.tex}
\input{section/5_conclusion.tex}

%\addtolength{\textheight}{-12cm}

%%%%%%%%%%%%%%%%%%%%%%%%%%%%%%%%%%%%%%%%%%%%%%%%%%%%%%%%%%%%%%%%%%%%%%%%%%%%%%%%

%%%%%%%%%%%%%%%%%%%%%%%%%%%%%%%%%%%%%%%%%%%%%%%%%%%%%%%%%%%%%%%%%%%%%%%%%%%%%%%%

%%%%%%%%%%%%%%%%%%%%%%%%%%%%%%%%%%%%%%%%%%%%%%%%%%%%%%%%%%%%%%%%%%%%%%%%%%%%%%%%
%\section*{Appendix}

%\section*{Acknowledgment}

%%%%%%%%%%%%%%%%%%%%%%%%%%%%%%%%%%%%%%%%%%%%%%%%%%%%%%%%%%%%%%%%%%%%%%%%%%%%%%%%

%References are important to the reader; therefore, each citation must be complete and correct. If at all possible, references should be commonly available publications.

\printbibliography
% \bibliographystyle{IEEEtran}
% \bibliography{Writing-paper_GlobalLocalization}

\end{document}

%% file: section/1_introduction.tex
\section{Introduction}

%Localization is 
%for stable and reliable navigation.
Global localization is a fundamental ability of mobile robots
where a pose of a robot is estimated given a sensor observation
without prior information of its ego-location.
It is used, e.g., for wake-up in a prior map, and recovery from localization failures.
%Various localization methods using different sensors are being developed,
%%These methods include using range sensors like
%such as monocular/stereo/RGB-D cameras, laser range finders (LRFs), etc.
%along with monocular cameras.
%While range and depth sensors are generally more expensive than monocular cameras,
%the latter, being low-cost and lightweight,
%are more suitable for mobile robots.
Monocular cameras are a popular choice as a sensor
for their nature of low cost and light weight,
as well as the rich visual information available.

%Among localization methods using a monocular camera,
One of the most popular approaches to
visual localization is to use a feature map
built by %local feature-based 
visual SLAM
(Simultaneous Localization And Mapping) systems,
such as ORB-SLAM series \cite{Mur-Artal2015,Mur-Artal2017,Campos2021}.
These methods calculate corresponding points between images and maps
based on the feature similarity, and then compute the camera pose.
Local features capture the distribution of local pixel values
viewed from a certain viewpoint;
therefore, changes in the viewpoint and camera resolution
during map generation and localization lead to inaccurate
correspondence matching and localization.

To address these challenges,
some studies proposed to use semantic objects as landmarks,
and image-based object detection as observation \cite{Nicholson2019a,Yang2019b}.
Particularly, deep learning-based object detectors are usually used
for their robustness to viewpoint and resolution changes.
Leveraging the characteristics,
object-based localization has been shown to be robust to those changes \cite{Zins2022,Wu2023a}.
Object-based maps also provide rich semantic information
that can be exploited in, e.g., manipulation \cite{Wu2023a}
and high-level task planning \cite{Kawasaki2021}.
% This part should be improved

Relocalization in object-based maps has not been actively tackled
in the existing work.
Most methods implement a simple iterative correspondence matching
via Random Sample Consensus (RANSAC) \cite{Fischler1981}
where all possible pairs of a landmark and an observation
with the same object class are treated as correspondence candidates
\cite{Zins2022,Zins2022a}.
This method is not feasible when the number of landmarks is large
due to exponential growth of the number of candidates.
%as the number of candidates increases exponentially.
Moreover, the existing method treats all the candidates equally.
%in the iteration.
In practice, there may be candidates that are almost certainly inliers
based on appearance.
Those should be favored more in the iterative sampling procedure
for more efficient calculation.

\begin{figure}[tb]
	\centering
	\includegraphics[width=\hsize]{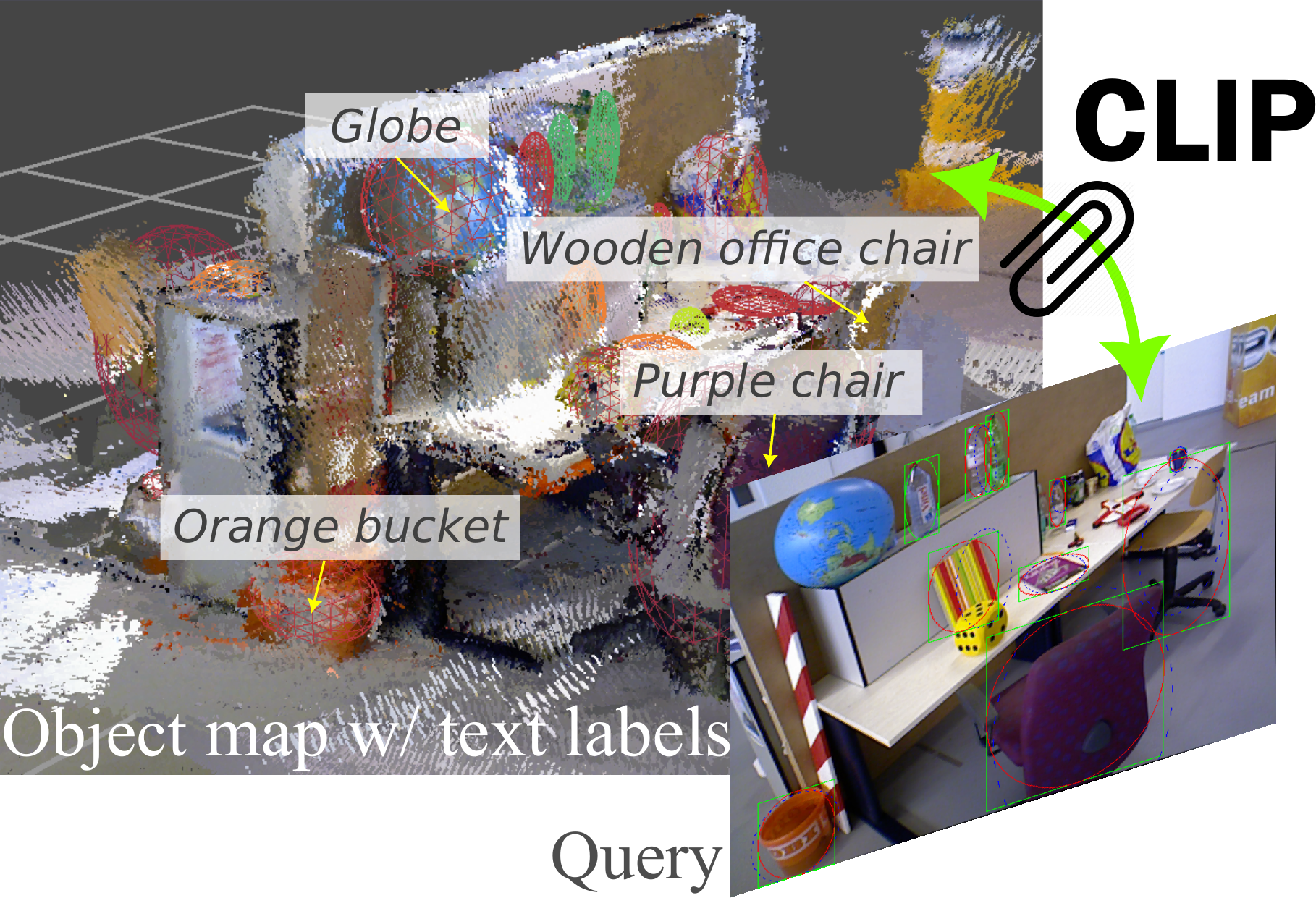}
	\vspace{-15pt}
	\caption{In conventional global localization in object-based maps,
		only class information is utilized for constructing  a
		correspondence candidate set, leading to exponential growth
		of the number as the landmark increases.
		In this paper, we propose to assign a natural language description
		as a label to each landmark, and match them with
		visual observation using a Vision Language Model (VLM).
		It enables more efficient correspondence matching
		leveraging the fine-grained information given by the text labels.}
	\label{fig:top_image}
\end{figure}

With the aforementioned discussions in mind,
in this work, we present a novel method of
landmark association for global localization in object-based maps.
First, we propose a method of multi-modal landmark association
leveraging an off-the-shelf vision and language model (VLM),
namely CLIP \cite{Radford2021}.
We assign each landmark a text label describing
its appearance and match them with visual observations
via the multi-modal embedding of CLIP. %(see Fig. \ref{fig:top_image}).
Utilizing rich information given by natural language labels,
the proposed method is able to establish
a correspondence candidate set with
only landmark-observation pairs that share similar concepts,
leading to fewer outliers and thus more efficient inlier extraction.
Second, we introduce an efficient iterative inlier extraction based on
Progressive Sample Consensus (PROSAC) \cite{Chum2005},
where correspondence candidates more likely to be inliers
are sampled more frequently.
We use the similarity score of the text and vision embeddings from CLIP
as a sampling weight for a correspondence candidate.

The main contributions of this work are as follows:
\begin{itemize}
	\item We propose a method to associate object landmarks with image observations by utilizing natural language labels and a VLM for efficient global localization.
	\item We introduce an efficient sampling method inspired by PROSAC \cite{Chum2005} using
	      text-image similarity of correspondence candidates as a score
	      for efficient computation.
\end{itemize}

%% file: section/2_related_work.tex
\section{Related Work}

\subsection{Object-based SLAM and relocalization}

Object-based SLAM is a subfield of
SLAM where a landmark is represented as
an object instance
in a form of CAD models \cite{Salas-Moreno2013},
%clusters of points, \cite{},
or primitive shapes (cubes \cite{Yang2019b},
planes \cite{Yang2019a},
quadrics \cite{Nicholson2019a,Zins2022,Zins2022a,ZiweiLiao2022},
superquadrics \cite{Han2022}, etc.).
It has been claimed that the object-based SLAM has some advantages
over feature-based visual SLAM methods,
such as rich semantic information of objects, and
robustness against view point changes \cite{Zins2022},
Especially, quadric-based object mapping has
been actively studied for
its compact representation and
a feature that projections of quadrics
onto an image plane can be efficiently calculated.

Although such object-based SLAM methods have been studied,
object-based relocalization is relatively under-explored.
% One of the reasons may be lack of discriminative
% information given by the objects.
% Most of ellipsoid-based SLAM methods provide
% approximated shape and size, 3D pose,
% and the object category of the landmarks.
% In global localization where prior information about
% the camera pose is not given,
As the object-based map lacks discriminative information
and only provides the object category, the size, and so on,
it has to rely on iterative methods such as RANSAC
to identify a set of inlier correspondences
from all possible matching candidates assigned with the same category,
which is not scalable to larger maps with more landmarks.

OA-SLAM by Zins et al. \cite{Zins2022} combines a feature-based SLAM
(ORB-SLAM2 \cite{Mur-Artal2017})
and the object-based landmark representation
for relocalization more robust to viewpoint changes.
However, the feature information is not exploited
in the object-based relocalization and
simple RANSAC is used.
Zins et al. \cite{Zins2022a} also proposed a camera pose estimation based on
a level-set metric that quantifies the alignment of the ellipses and
the projections of corresponding ellipsoids.
This, however, also depends on pose initialization by RANSAC.
Wu et al. \cite{Wu2023a} proposed a graph-based scene matching for
relocalization in object-centric maps.

Unlike the existing work,
we use natural language labeling and rich multi-modal information from CLIP
to reinforce the discriminativity of object landmarks and thus
to improve the efficiency of object-based global localization.

\subsection{Vision and Language Models}

Driven by the recent progress of
Transformer \cite{Zhai2021a}-based networks,
so-called \textit{foundation models} have been proposed
in a variety of tasks
\cite{Brown2020,Brohan2023}.
% models trained with enormous amount of data are proposed, 
Vision Language Models (VLMs) are a type of
models pre-trained with large-scale image-text pairs
so that they can be applied to visual and
linguistic inference tasks.
CLIP \cite{Radford2021} is a representative VLM
that can ground visual concepts
to natural language instructions.
%It allows zero-shot fine-grained image classification
%conditioned by qualitative natural language information.
CLIP has been applied to various
applications with open vocabulary scene inference such as
visual navigation \cite{Shah2022,Gadre2023},
visual place recognition \cite{Mirjalili2023},
%open vocabulary scene inference, 
etc.

The work by Mirjalili et al. \cite{Mirjalili2023} is
the most related to our work in a way that
they use a VLM for visual localization.
In \cite{Mirjalili2023}, a per-image descriptor is constructed
exploiting CLIP and GPT-3 and conduct image retrieval
from pre-built image database.
The proposed method differs in that it is specifically tailored for
localization in object-based maps by
directly associating the visual observations with the landmarks.

\subsection{Random sampling strategies}

Random Sample Consensus (RANSAC) \cite{Fischler1981} is
a commonly used model estimator for robustly estimating
model parameters using a dataset that contains
a certain percentage of outliers
by iterating a procedure of data sampling, parameter estimation,
and verification.
In the context of correspondence matching,
the dataset is a set of correspondence hypotheses.
The original RANSAC uniformly samples candidates in each iteration.
Progressive Sample Consensus (PROSAC) \cite{Chum2005}
is a variant of RANSAC where the quality of each datum
is measured, and the data are sampled according to the quality score
to favor more promising samples.
Although some other sampling strategies have been proposed,
e.g., MAGSAC++ \cite{Barath2020},
we adopt PROSAC for its simplicity of implementation and the suitability
to our task where correspondence candidates are naturally
assigned with a similarity score between image and text embeddings.

%% file: section/3_proposed_method.tex
\section{Proposed Method}
\label{sec:proposed_method}
\begin{figure*}[tb]
	\centering
	\includegraphics[width=0.94\hsize]{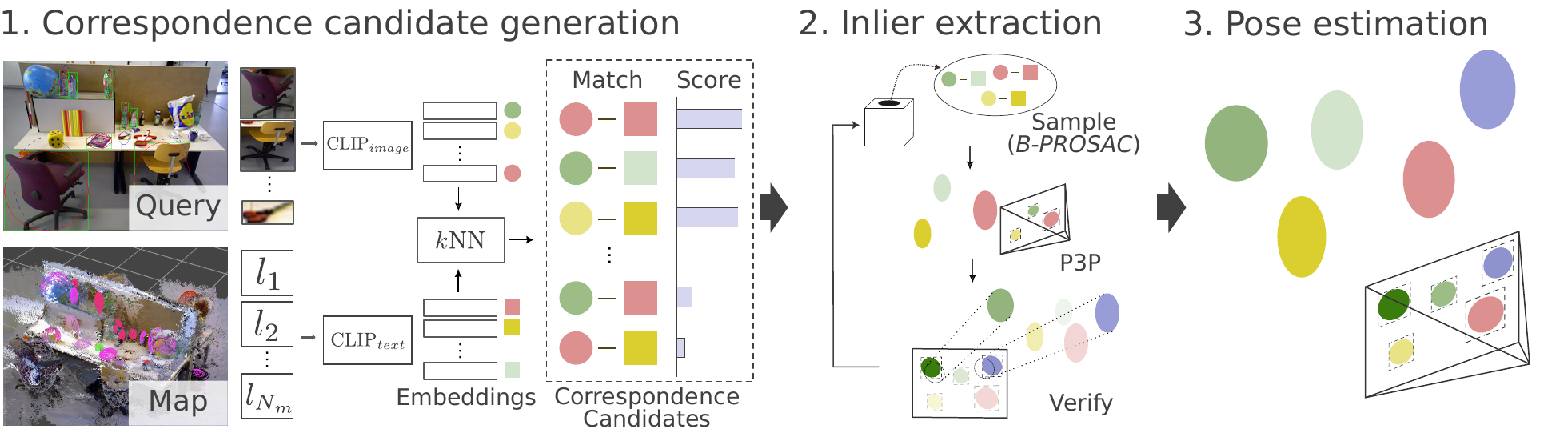}
	\vspace{-8pt}
	\caption{Overview of the proposed method. We assume that object landmarks in the map are
		labeled by natural language descriptions. In addition, for a query image, object locations and classes
		are given by a general object detector (here we use YOLOv8). Given those data, a camera pose is
		estimated in the following steps:
		1) The query objects and the landmark labels are embedded to a common feature space
		using image and text encoders of CLIP \cite{Radford2021}, respectively. For each image embedding,
		$k$ nearest neighboring text embeddings are retrieved, and correspondence candidates are generated.
		2) Inlier correspondences are estimated through B-PROSAC described in Sec. \ref{sec:proposed_method_inlier_extraction}.
		It consists of sampling three candidates considering the matching score, pose estimation via P4P, and
		pose verification.
		The final pose is given as the one with the best score after a certain number of iterations.
		3) Optionally, the pose can be further refined by methods, e.g., \cite{Zins2022a}.
	}
	\label{fig:overview}
\end{figure*}

\subsection{Problem formulation}

We assume an object map represented as
ellipsoidal landmarks with text labels
in a form of natural language descriptions.
Text labels can be given by any methods
such as manual input, voice dictation,
or image captioning, e.g., \cite{Wang2022d}.
As a query, a single image is given.
Bounding boxes of objects in the image
are detected using an arbitrary object detector
and used as observations.

Formally, a map $\mathcal{M}$ consists of
a set of $N_m$ object landmarks $\{\mathbf{Q}^{*}_i, l_i\}_{i=1}^{N_m}$,
where $\mathbf{Q}^{*}_i$ and $l_i$ denote
a dual form of a quadric representing the landmark,
and a text label assigned to the landmark, respectively.
A query image is denoted as $I$,
and detected object bounding boxes are
denoted as $\{\mathbf{b}_j\}_{j=1}^{N_o}$,
where $N_o$ is the number of detected objects.

%In our experiment, we use YOLOv8
%\footnote{\url{https://github.com/ultralytics/ultralytics}}.
\subsection{Overview of the algorithm}
\label{sec:proposed_method_overview}

The overview of the method is shown in Fig. \ref{fig:overview},
and pseudo-code is shown in Algorithm \ref{alg_clip-loc}.
The process of the proposed method consists of three steps:
1) correspondence candidate generation,
2) inlier correspondence extraction, and
3) pose estimation.
In correspondence candidate generation,
observation-landmark matches are searched for
based on the similarity of observations
and text-labeled landmarks given by CLIP.
Inlier correspondences are then extracted
via PROSAC \cite{Chum2005}-like weighted sampling,
and verification based on overlap of
observations and the projected corresponding landmarks.
The solution is given as the one with the best score.
Finally, the camera pose can be further refined
using the best correspondences.
% Finally, the camera pose is calculated using
% the best correspondences.

\begin{algorithm}[tb]
	\caption{CLIP-Loc}
	\label{alg_clip-loc}
	\begin{algorithmic}[1]
		\REQUIRE Map instance set $\mathcal{M}=\left\{Q_i, l_i\right\}^{N_m}_{i=1}$, query instance set $\mathcal{Q}=\left\{q_j\right\}^{N_q}_{j=1}$, the number of loops $N$, the number of nearest neighbors $k$ to search for each observation
		\ENSURE Camera pose $\mathbf{p}$, correspondence set $\mathcal{C}^{*}$
		\STATE \# Text and image embedding via CLIP
		\STATE $\mathcal{F}_m, \mathcal{F}_q=\text{CLIP}(\mathcal{M}, \mathcal{Q})$
		\STATE \# Correspondence candidate set
		\STATE $\mathcal{F}^{*}_m=\text{kNN}(\mathcal{F}_m, \mathcal{F}_q, k)$
		\STATE $\mathcal{C}=\mathcal{F}^{*}_m \times \mathcal{F}_q$
		\STATE Sort $\mathcal{C}$ by the method described in Sec. \ref{sec:sampling_method}
		\STATE
		\STATE \# PROSAC loop
		\STATE $max\_score=-1$
		\FOR{i in \text{range}(N)}
		\STATE \# Get a sample set $\mathbf{s}$ by PROSAC sampling \cite{Chum2005}
		\STATE $\mathbf{s}=\text{prosac\_sample}(\mathcal{C})$
		\STATE \# Calculate a camera pose via P3P
		\STATE $\mathbf{p}_{tmp}=\text{P3P}(\mathbf{s})$
		\STATE \# Calculate $score$, and estimate potential correspondence set $\hat{\mathcal{C}}$
		\STATE $score, \hat{\mathcal{C}}=\text{verify}(\mathbf{p}_{tmp}, \mathcal{M}, \mathcal{Q})$
		\IF{$score > max\_score$}
		\STATE $\mathcal{C}^{\ast} \leftarrow \hat{\mathcal{C}}$
		\STATE $max\_score \leftarrow score$
		\STATE $\mathbf{p} \leftarrow\mathbf{p}_{tmp} $
		\ENDIF
		\ENDFOR
		% \STATE $p\leftarrow$calculate\_pose$\left(\mathcal{C}^{\ast}\right)$
		\RETURN $\mathbf{p}$, $\mathcal{C}^{*}$
	\end{algorithmic}
\end{algorithm}

\subsection{CLIP-based correspondence candidate generation}
\label{sec:proposed_method_candidate_generation}

%\subsubsection{General algorithm of generating correspondence candidates}

As an offline pre-processing,
a text label for each map landmark is
encoded by the text encoder of CLIP
and stored as a set of embeddings
$\mathcal{F}^m=\{f^{m}_{i}\}^{N_m}_{i=1}$, where
\begin{equation}
	f^{m}_i = \text{CLIP}_{text}\left(l_i\right).
\end{equation}
When an image $I$ is given,
object regions are extracted by the object detector.
A visual embedding $f_j$ for an image cropped by
bounding box $\mathbf{b}_j$ is extracted by CLIP image encoder
$\text{CLIP}_{image}\left(\cdot\right)$:
\begin{equation}
	f^{o}_j = \text{CLIP}_{image}\left(\text{crop}\left(\mathbf{b}_j, I\right)\right),
\end{equation}
where $\text{crop}\left(\cdot, \cdot\right)$ denotes
a function to return a given image cropped with a given bounding box.

After getting the image embeddings,
for each image embedding $f^o_j$,
we retrieve $k$ nearest neighboring
text embeddings in the embedding space:
\begin{equation}
	\mathcal{F}^{m*}_{j} = k\text{NN}\left(\mathcal{F}^m, f^o_j, k\right).
\end{equation}
A set of correspondence candidates that has $\mathbf{b}_j$ as
an observation is given as:
\begin{equation}
	\mathcal{C}_{j} = \{f^o_j\} \times \mathcal{F}^{m*}_{j}.
\end{equation}
The whole set of correspondence candidates is thus given as:
\begin{equation}
	\mathcal{C} = \bigcup_{j=1}^{N_o} \mathcal{C}_{j}.
\end{equation}

\subsection{Efficient inlier extraction}
\label{sec:proposed_method_inlier_extraction}

\subsubsection{Sampling method}
\label{sec:sampling_method}

Intuitively, the higher the similarity score
between an observation and a landmark,
the more likely that they are an inlier correspondence.
PROSAC \cite{Chum2005} is based on such an intuition.
In PROSAC, the correspondence candidates are
sorted by some measure of the quality of matching.
Correspondence hypotheses are sampled from
only promising candidates in the earlier iterations,
and the sampling strategy gradually shifts
towards the original RANSAC as the iteration progresses.
For details, the readers are referred to \cite{Chum2005}.

\begin{figure}[tb]
	\centering
	\includegraphics[width=0.75\hsize]{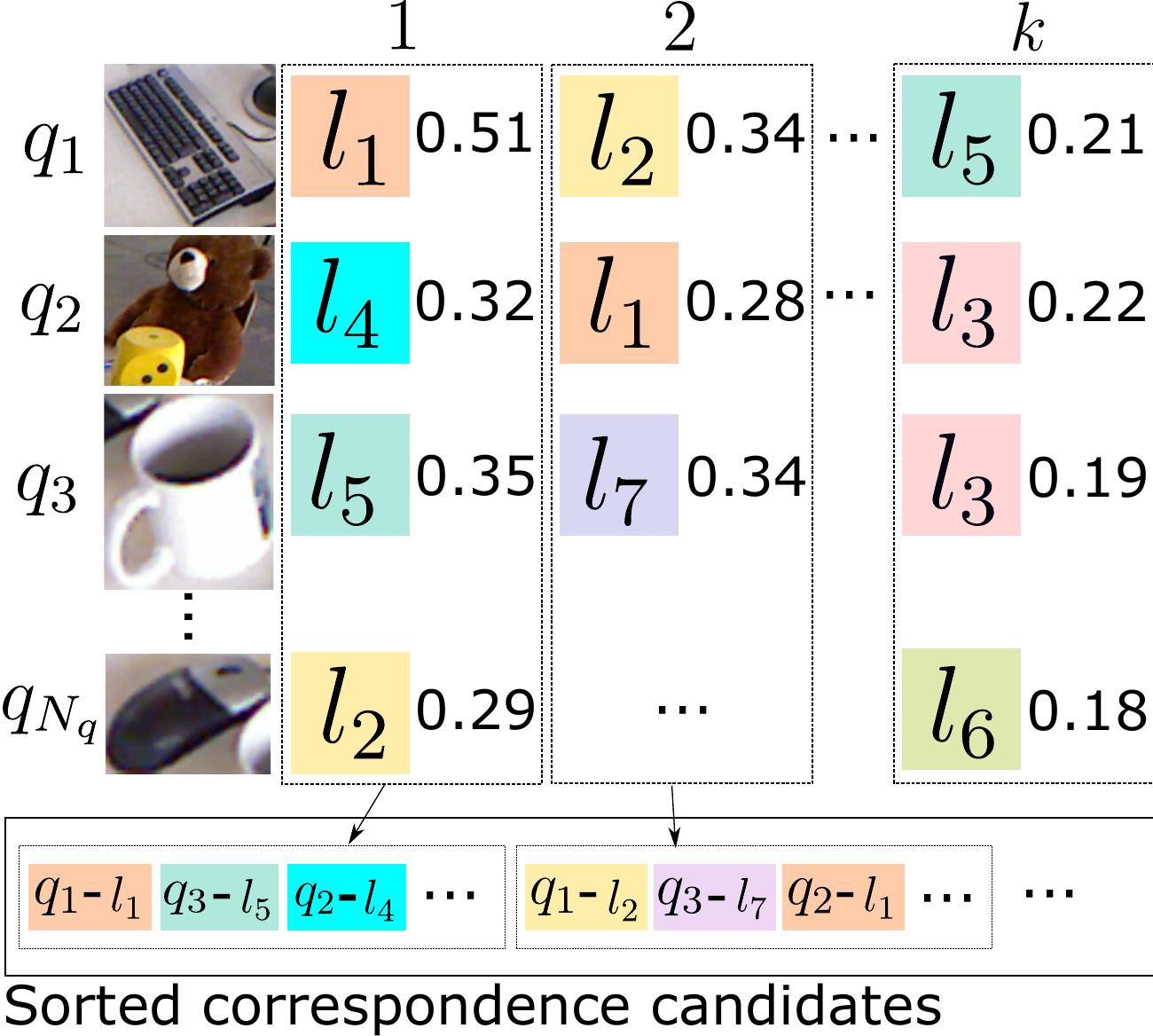}
	\vspace{-7pt}
	\caption{Sorting in the proposed method. Top 1 nearest landmarks
		for the observations are first sorted by the score
		and stored in an ordered correspondence candidate list.
		Top 2 landmarks are then sorted and concatenated to
		the list, and so forth.}
	\label{fig:prosac_sort}
\end{figure}

In our problem,
the matching score can be naturally acquired as
a cosine similarity between the image and the text embeddings,
so PROSAC is suitable to the problem.
%Here, we briefly summarize the algorithm \cite{Chum2005}.
% Utilizing the similarity score between
% the observation and the text-labeled landmark,
% we can further improve the efficiency of
% the inlier extraction.
%In the original PROSAC,
%the correspondence candidates are sorted by the matching score,
%and only candidates with high scores are sampled in earlier iterations
%while gradually expanding the sampling range, getting closer to
%the ordinary RANSAC.
%and those with higher scores are sampled more intensively.
We, however, found that sorting simply by the cosine similarity
led to imbalanced samples biased to
detections with larger regions of interest (ROI).
We, therefore, sort the candidates such that
the observations included in the candidates are balanced.
The nearest landmarks for the observations are first
sorted in the descending order of the similarity scores.
The second-nearest landmarks are sorted next and stored,
and so forth
% まず各画像検出の最近傍ランドマークをスコア順にソート．
% 次に各画像検出のsecond nearestのランドマークをスコア順にソート
% 以下k回繰り返し
(see Fig. \ref{fig:prosac_sort}).
After sorting, the inliers are estimated in
the same procedure as PROSAC.
We refer to the modified version of PROSAC
as B-PROSAC (\underline{B}alanced-PROSAC).

\subsubsection{Pose verification}
In each iteration of PROSAC/B-PROSAC,
a camera pose is calculated using the sampled correspondence candidates
and verified based on consistency of other observations with
potentially corresponding landmarks.
Using the three sampled correspondences,
up to four camera poses are calculated by solving
the Perspective-3-Point (P3P) problem.
The most likely pose is chosen by
evaluating the overlaps between the observations
and the projected ellipses of the corresponding ellipsoid landmarks.

Once the pose is determined,
correspondences for observations
not in the samples are estimated
by projecting potentially corresponding landmarks
onto the image from the pose.
% For the verification,
% landmarks that have the same object class with
% the observation are used as potential correspondences.
For an observation $\mathbf{b}_j$,
a landmark is considered as a correspondence
if the IoU of an ellipsoid fit to the observation bounding box $\mathbf{b}_j$
and the projection of the landmark is larger than a threshold and
that of the ellipsoid and the projection of any other potentially corresponding landmarks.
The pose is scored with the sum of the IoUs for the observations.

After the pre-determined number of iterations,
the final pose is given as the one with the best score.
Optionally, the pose can be further refined
by methods, e.g., \cite{Zins2022a}.

\noindent
\textbf{Landmarks used for pose verification}
\label{sec:proposed_method_pose_verification}
% \subsubsection{Hybrid matching strategy}
%Ideally, the correspondence candidates for all detections
%include inliers which can be extracted in
%the downstream process.
We found that CLIP tends to wrongly estimate
the matching for small observations. %image detections,
%especially small ones.
%with low detection confidence.
% This means that the $k$ nearest correspondence candidates
% for low confidence detections do not include the true correspondence.
%At the same time,
Using those erroneous candidates in verification
leads to unreliable results.
Therefore, we introduce a ``hybrid'' matching strategy
where %only high quality detections are used
%CLIP-based correspondence candidates are used
%as correspondence candidates via CLIP embedding,
%for pose calculation
% and the rest are used for verification
both CLIP-based correspondences
%while the calculated pose is verified
and those based on the object class information
are used for pose verification.
By this way, we can reduce outlier candidates leveraging CLIP,
while still utilizing the class information,
which is more reliable for small objects.

% \subsection{Pose estimation}

% The overall algorithm of the proposed method is shown in
% Algorithm \ref{alg_clip-loc}.

%% file: section/4_experiment.tex
\section{Experiments}
\label{sec:experiments}

\subsection{Setup}
\label{sec:experiments_setup}

\subsubsection{Datasets}
\label{sec:experiments_setup_datasets}
We evaluated our method with two datasets.
The first one is TUM dataset \cite{Sturm2012}.
We use \textit{fr2/desk}
and \textit{fr3/long\_office\_household} sequences
for evaluation.
The sequences include 60 and 52 frames, respectively,
extracted by sampling every 50 frames from
the original image sets.
%The second one is a middle-scale data
%collected by our own.
%We collected RGB-D sequences and odometry
%using Toyota Human Support Robot (HSR) \cite{Yamamoto2019}.
%We then used RTAB-Map \cite{Labbe2018} to
%construct a point cloud map and
%to yield pseudo-ground-truth camera poses
%for the test frames.
For both datasets, we reconstructed 3D point cloud
of the scenes, manually fitted an ellipsoid,
and assigned a text label to each object using a self-made GUI tool
to build consistent maps.

\subsubsection{Implementation details}
\label{sec:experiments_setup_impl_details}

To acquire image observations,
we used YOLOv8-x model\protect\footnotemark
trained with COCO dataset \cite{Lin2014}.
\footnotetext{\url{https://github.com/ultralytics/ultralytics}}
We saved the predictions of the model with confidence higher than 0.1
for the test images offline
and used them in all experiments for
fair comparison among the different algorithms and trials.
As an implementation of CLIP \cite{Radford2021},
we used ViT-L/14 model shared by OpenAI \protect\footnotemark.
\footnotetext{\url{https://github.com/openai/CLIP}}

% \subsubsection{Variants of the proposed method}
All algorithms were implemented in Python 3
mainly for integration with
official implementations of YOLOv8 and CLIP,
and without any parallelization.
Although the implementation in pure Python is not optimal
from a practical point of view,
the algorithm is parallelizable and
the computational efficiency can be further improved.
Here, we focused on evaluation of
the efficiency of the proposed algorithm
relative to the baselines.
In addition, we report the results
without pose refinement using the found correspondences
for all methods.
The experiments were conducted on
a computer with an RTX 4090
and Intel Core i9.

For the proposed correspondence candidate generation,
the parameter $k$ (the number of nearest text embeddings
searched for each observation) is empirically set to 3.
%by empirical parameter search

\subsubsection{Baselines}

We compare different combinations of
matching types for generating correspondence candidates,
and algorithms for inlier extraction.

\noindent
\textbf{Matching types}
We compare three matching types, namely \textit{class}, \textit{clip}, and \textit{hybrid}.
\textit{class}: Pairs of an observation and a landmark with the same
category are considered correspondence candidates and the same set is used for pose verification.
\textit{clip}: Correspondence candidates generated by the method described in Sec. \ref{sec:proposed_method_candidate_generation}
are used for both pose calculation and pose verification.
\textit{hybrid}: The same correspondence candidates as \textit{clip} are used for pose calculation, while
those of both \textit{clip} and \textit{class} are used for pose verification, as described in Sec. \ref{sec:proposed_method_pose_verification}.

% \begin{itemize}
% 	\item \textit{class}: Pairs of an observation and a landmark with the same
% 	      category are considered correspondence candidates, and the same set is used for pose verification.
% 	\item \textit{clip}: Correspondence candidates generated by the method described in Sec. \ref{sec:proposed_method_candidate_generation}
% 	      are used for both pose calculation and pose verification.
% 	\item \textit{hybrid}: The same correspondence candidates as \textit{clip} are used for pose calculation, while
% 	      those of \textit{class} are used for pose verification, as described in Sec. \ref{sec:proposed_method_pose_verification}.
% \end{itemize}

\noindent
\textbf{Algorithms}
We compare four algorithms, namely \textit{bf}, \textit{ransac}, \textit{prosac}, and \textit{b-prosac}.
\textit{bf}: Brute force search over all possible combinations of candidates.
\textit{ransac}: Ordinary RANSAC \cite{Fischler1981}.
\textit{prosac}: PROSAC \cite{Chum2005} considering only the score.
\textit{b-prosac}: B-PROSAC described in \ref{sec:sampling_method} considering the balance among the observations.
%\begin{itemize}
%	\item \textit{bf}: Brute force search over all possible combinations of candidates.
%	\item \textit{ransac}: Ordinary RANSAC \cite{Fischler1981}.
%	\item \textit{prosac}: PROSAC \cite{Chum2005} considering only the score.
%	\item \textit{b-prosac}: B-PROSAC described in \ref{sec:sampling_method} considering
%	      the balance among the observations.
%\end{itemize}

We hereafter denote the methods by
a notation of ``\{matching type\}\_\{algorithm\}'',
such as \textit{hybrid\_b-prosac}.
Note that \textit{class\_bf} is used in the official implementation
\protect\footnotemark
of OA-SLAM \cite{Zins2022}, and \textit{class\_ransac}
is used in major object-based SLAM systems
such as \cite{Zins2022a}.
\footnotetext{\url{https://gitlab.inria.fr/tangram/oa-slam}}

\subsubsection{Evaluation metrics}

We evaluate the results with two metrics:
\textit{success rate}, calculated as
the ratio of test queries for which
a pose is estimated with an error less than a threshold
against all test queries, and
\textit{translation error},
calculated as the Euclidean distance
between the estimated and the groundtruth poses.

\subsection{Comparative studies}

\begin{table*}[tb]
	\centering
	\caption{Comparison with the baselines}
	\label{table:comparative_result}
	\begin{threeparttable}
		\begin{tabular}{ c c c c c c c c }
			\toprule
			                        &          & \multicolumn{3}{c}{\textit{fr2}} & \multicolumn{3}{c}{\textit{fr3}}                                                                                                                             \\
			\cmidrule(lr){3-5}
			\cmidrule(lr){6-8}
			                        &          & Success rate $\uparrow$          & Trans. error [m]$\downarrow$     & Time [sec] $\downarrow$      & Success rate $\uparrow$      & Trans. error [m]$\downarrow$ & Time [sec] $\downarrow$      \\
			\midrule
			\multirow{2}{*}{class}  & bf       & $0.75\pm0.0$                     & $0.97\pm0.0$                     & $0.264\pm0.22$               & $0.8\pm0.0$                  & $0.749\pm0.0$                & $245\pm569$                  \\
			\cdashlinelr{2-8}
			                        & ransac   & $0.75\pm0.0149$                  & $0.989\pm0.0368$                 & $\mathbf{0.311\pm0.0433}$    & $0.684\pm0.0344$             & $1.02\pm0.0795$              & $0.563\pm0.168$              \\

			\midrule

			\multirow{4}{*}{clip}   & bf       & $0.817\pm0.0$                    & $0.802\pm0.0$                    & $5.26\pm1.35$                & $0.68\pm0.0$                 & $1.02\pm0.0$                 & $5.37\pm1.57$                \\
			\cdashlinelr{2-8}
			                        & ransac   & $0.743\pm0.0455$                 & $0.887\pm0.0829$                 & $0.515\pm0.04$               & $0.644\pm0.0463$             & $1.13\pm0.147$               & \underline{$0.525\pm0.0668$} \\
			                        & prosac   & $0.79\pm0.0554$                  & $0.819\pm0.121$                  & \underline{$0.497\pm0.0427$} & $0.632\pm0.0371$             & $1.14\pm0.106$               & $\mathbf{0.507\pm0.0684}$    \\
			                        & b-prosac & $0.757\pm0.0343$                 & $0.878\pm0.0945$                 & $0.498\pm0.0426$             & $0.68\pm0.0179$              & $1.08\pm0.0729$              & $\mathbf{0.507\pm0.07}$      \\

			\midrule

			\multirow{4}{*}{hybrid} & bf       & $0.85\pm0.0$                     & $0.682\pm0.0$                    & $6.38\pm1.66$                & $0.76\pm0.0$                 & $0.796\pm0.0$                & $8.77\pm2.83$                \\
			\cdashlinelr{2-8}
			                        & ransac   & $0.773\pm0.0226$                 & $0.827\pm0.0345$                 & $0.649\pm0.0682$             & $0.668\pm0.0412$             & \underline{$0.983\pm0.0744$} & $0.943\pm0.222$              \\
			                        & prosac   & $\mathbf{0.817\pm0.0394}$        & $\mathbf{0.701\pm0.0475}$        & $0.624\pm0.0681$             & \underline{$0.708\pm0.0299$} & $\mathbf{0.927\pm0.0589}$    & $0.916\pm0.223$              \\
			                        & b-prosac & \underline{$0.793\pm0.017$}      & \underline{$0.718\pm0.0397$}     & $0.626\pm0.0696$             & $\mathbf{0.72\pm0.0283}$     & \underline{$0.983\pm0.0638$} & $0.926\pm0.226$              \\
			\bottomrule
		\end{tabular}
		\begin{tablenotes}[normal] %(default:normal)
			\item \textbf{Bold} and \underline{underline} denote the best and the second-best among the sampling base algorithms
			(\textit{ransac}, \textit{prosac}, \textit{b-prosac}), respectively.
			%The results of the brute-force algorithm
			%are shown as the upper bound of performance that can be achieved using each matching type.
		\end{tablenotes}

	\end{threeparttable}

\end{table*}

% \begin{figure*}[tb]
% 	\centering
% 	\subfloat[\textit{fr2} \label{fig:comparative_fr2}]{
% 		\includegraphics[width=0.40\hsize]{figs/rgbd_dataset_freiburg2_desk_class,clip,hybrid_ransac,prosac_N=50,100,300,500_k=3_N=500_k=3_roc}}
% 	\subfloat[\textit{fr3} \label{fig:comparative_fr3}]{
% 		\includegraphics[width=0.40\hsize]{figs/rgbd_dataset_freiburg3_long_office_household_class,clip,hybrid_ransac,prosac_N=50,100,300,500_k=3_N=500_k=3_roc}}
% 	\caption{Success rate with regard to error thresholds}
% 	\label{fig:comparative}
% \end{figure*}

\begin{figure}[tb]
	\centering
	\subfloat[\textit{fr2} \label{fig:comparative_fr2}]{
		\includegraphics[width=0.8\hsize]{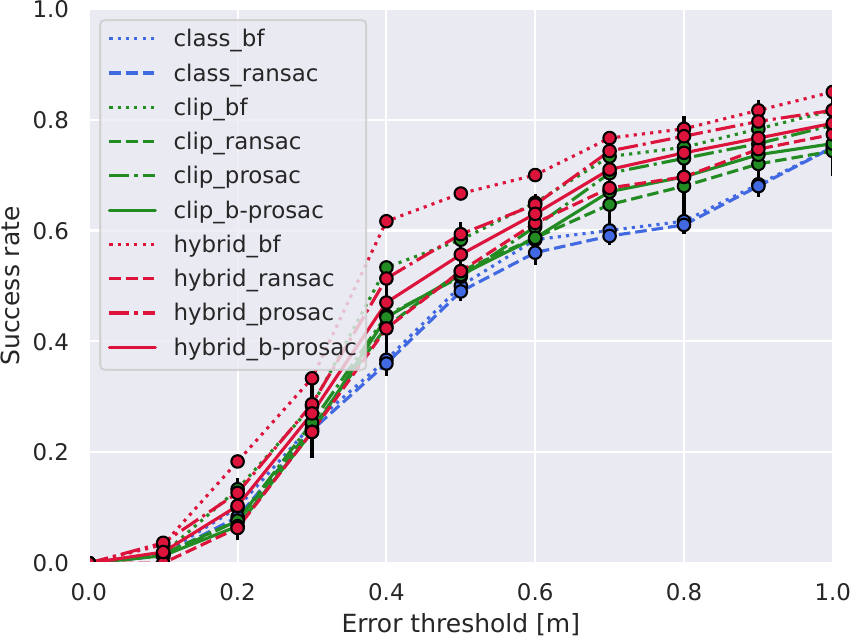}} \\
	\subfloat[\textit{fr3} \label{fig:comparative_fr3}]{
		\includegraphics[width=0.8\hsize]{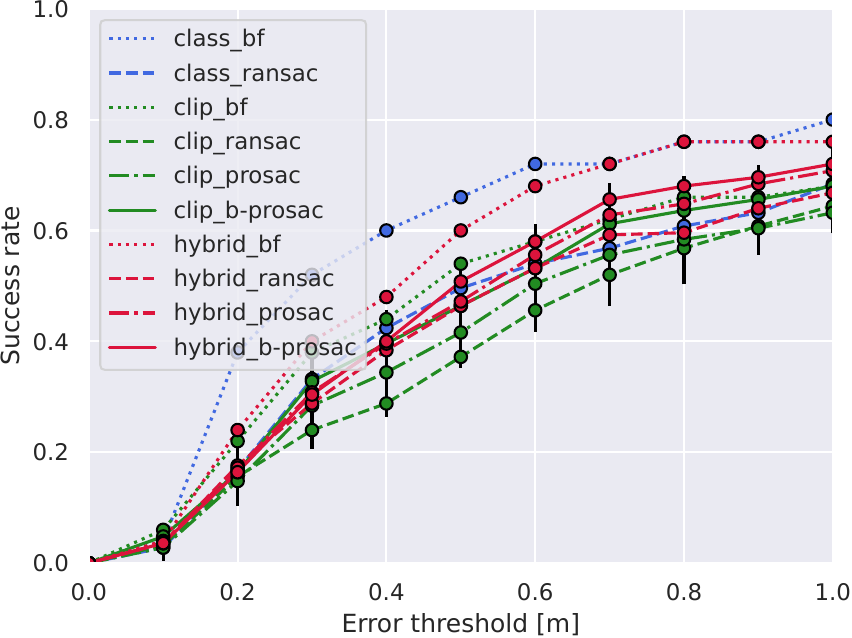}}
	\caption{Success rate with regard to error thresholds when the number of iteration is 500.
		The proposed methods (\textit{hybrid\_prosac}/\textit{hybrid\_b-prosac})
		outperformed the category-based baseline.}
	\label{fig:comparative}
\end{figure}

% \subsection{Parameter analysis}
% 
% The choice of algorithm (CLIP-Loc or normal RANSAC)
% and the optimal parameters such as the number of iterations
% are closely related.

%\begin{figure*}[tb]
%	\centering
%	\subfloat[\textit{fr2} \label{fig:param_analysis_n_iter_fr2}]{
%		\includegraphics[width=0.40\hsize]{figs/rgbd_dataset_freiburg2_desk_class,clip,hybrid_ransac,prosac_N=50,100,300,500_k=3__success rate_vs_N_k=3}}
%	\subfloat[\textit{fr3} \label{fig:param_analysis_n_iter_fr3}]{
%		\includegraphics[width=0.40\hsize]{figs/rgbd_dataset_freiburg3_long_office_household_class,clip,hybrid_ransac,prosac_N=50,100,300,500_k=3__success rate_vs_N_k=3}}
%	\caption{Success rate with different numbers of iterations. \textit{hybrid\_b-prosac}
%		with only 100 iterations outperformed \textit{hybrid\_ransac}
%		and \textit{class\_ransac} with more iterations,
%		showing the efficiency of the proposed method.}
%	\label{fig:param_analysis_n_iter}
%\end{figure*}
\begin{figure}[tb]
	\centering
	\subfloat[\textit{fr2} \label{fig:param_analysis_n_iter_fr2}]{
		\includegraphics[width=0.8\hsize]{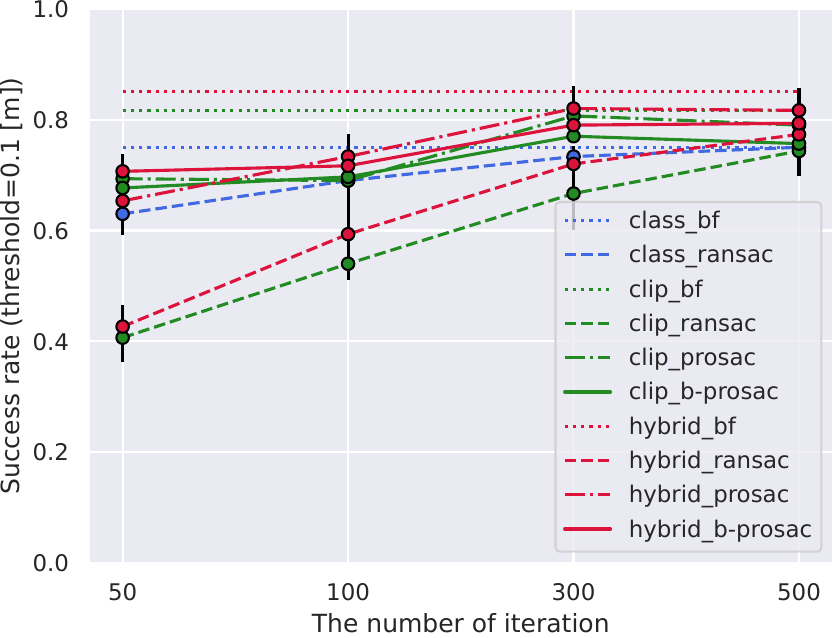}} \\
	\subfloat[\textit{fr3} \label{fig:param_analysis_n_iter_fr3}]{
		\includegraphics[width=0.8\hsize]{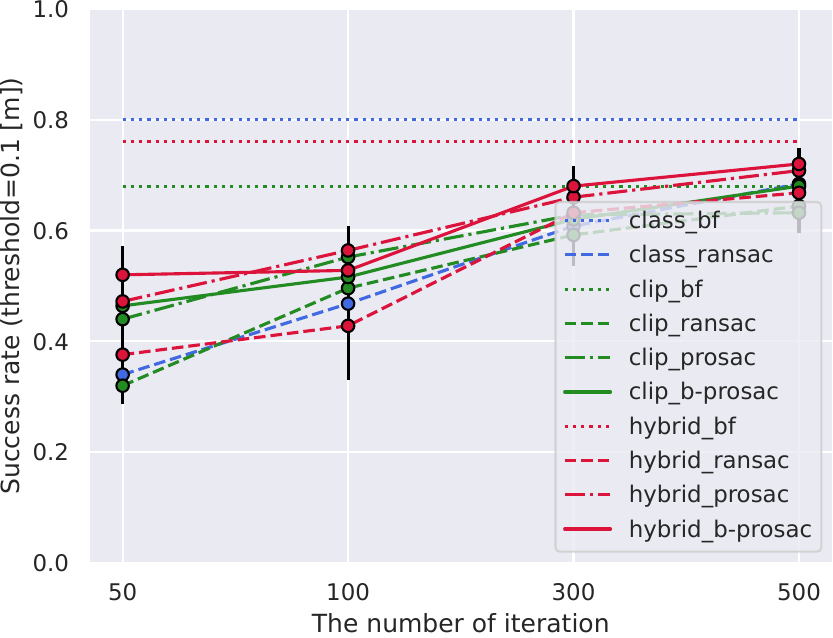}}
	\caption{Success rate with different numbers of iterations.
		\textit{hybrid\_prosac} and \textit{hybrid\_b-prosac}
		consistently outperformed the counterparts relying on
		category information and RANSAC.}
	\label{fig:param_analysis_n_iter}
\end{figure}

%We first evaluate the general performance of the proposed method.
The comparative results with the baselines with the number of iterations
$N=500$ are shown in
Table \ref{table:comparative_result}, and
Fig. \ref{fig:comparative} shows the success rate
over the sequences with respect to
translation error thresholds.
We report the average and standard deviation of
five trials for each method.

\noindent
\textbf{Comparison of matching types}
Hybrid matching with PROSAC/B-PROSAC
resulted in better performance
than the object class-based baselines
in both datasets.
%than other matching types with the same algorithms.
\textit{hybrid\_prosac} and \textit{hybrid\_b-prosac}
even outperformed \textit{class\_bf}.
%exhibiting efficiency and effectiveness of the proposed method.
% Interestingly, brute force search on \textit{hybrid}
% matching strategy resulted in better success rate and error
% compared to brute force with different matching strategies.
This is because \textit{hybrid} focuses only on
promising candidates and thus reduces the outlier ratio.
The proposed correspondence generation enables efficient
correspondence search leveraging the fine-grained information
of text labels and the VLM.
\textit{clip} resulted in worse performance than \textit{hybrid},
justifying the proposed matching strategy.

In \textit{fr2}, \textit{class\_bf} is a reasonable choice
because of the moderate number of landmarks,
but it soon became infeasible in \textit{fr3}
requiring more than 8 minutes on average.
On the other hand, the proposed matching strategy
makes the brute force algorithm
feasible also on \textit{fr3} with only a few seconds.

\noindent
\textbf{Comparison of algorithms}
% As expected, brute force search performed the best,
% but computationally inefficient
% depending on the number of objects.
B-PROSAC or PROSAC in combination with \textit{hybrid}
matching type consistently resulted in better performance
with more efficient computation
compared to RANSAC-based counterparts,
effectively exploiting the similarity scores
given by the proposed CLIP-based correspondence matching.
To take a closer look at the effect of PROSAC / B-PROSAC,
we compare the success rates with different numbers
of iterations, shown in Fig. \ref{fig:param_analysis_n_iter}.
\textit{hybrid\_b-prosac} resulted in better success rate
than other algorithms and matching types
even with fewer iterations,
exhibiting the efficiency of the proposed method.
% CLIPが低い→it lacks the landmarks for verification
%The effect of B-PROSAC to localization performance
%was not clear in the experiments.

\begin{figure}[tb]
	\centering
	\subfloat[\textit{fr2} \label{fig:qualitative_results_fr2}]{
		\includegraphics[width=0.49\hsize]{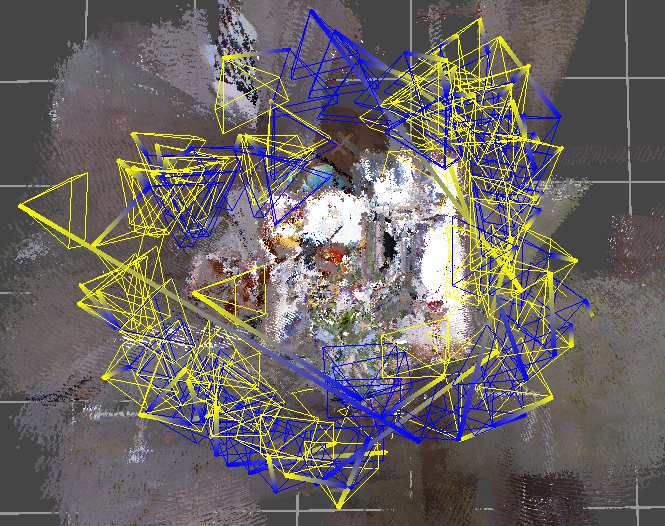}}
	\subfloat[\textit{fr3} \label{fig:qualitative_results_fr3}]{
		\includegraphics[width=0.49\hsize]{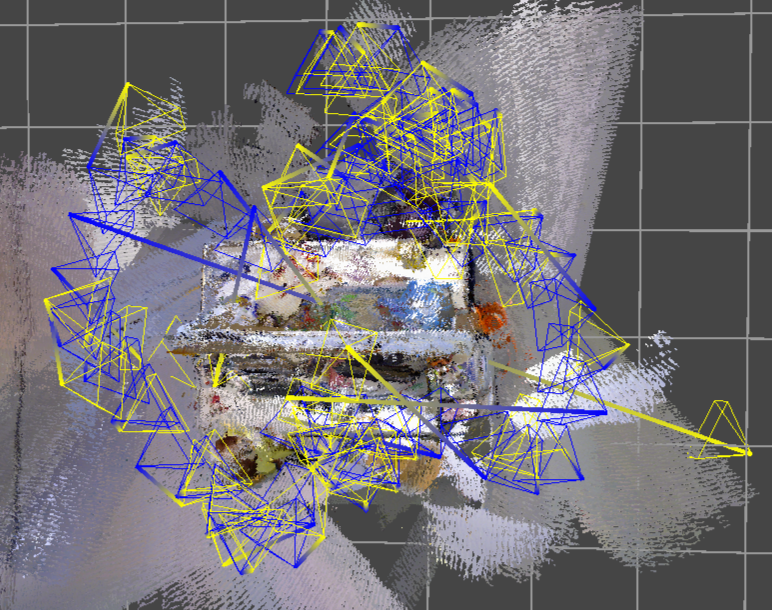}}
	\caption{Top view of the estimation results by
		\textit{hybrid\_b-prosac}.
		Estimatd frames and groundtruth frames are in yellow and blue, respectively,
		and an estimation and its groundtruth is connected by a line.
		Best viewed in color.}
	\label{fig:qualitative_results}
\end{figure}

Fig. \ref{fig:qualitative_results} shows qualitative results of
the proposed \textit{hybrid\_b-prosac} with
the iteration number of 500.

% \begin{table}[tb]
% 	\centering
% 	\caption{}
% 	\label{table:experiment_runtime_analysis}
% 
% 	\begin{tabular}{ c c c }
% 		\toprule
% 		             & RANSAC & \textbf{Proposed} \\
% 		\midrule
% 		\textit{fr2} & 10000  & 100               \\
% 		\textit{fr3} & -      &                   \\
% 		\bottomrule
% 	\end{tabular}
% \end{table}
% 
\subsection{Convergence of the algorithms}

%The time required for each method is shown in Table \ref{table:comparative_result}.
%Although \textit{hybrid} strategy resulted in slightly worse runtime,
%it

\begin{figure}[tb]
	\centering
	\subfloat[\textit{fr2} \label{fig:runtime_analysis_fr2}]{
		\includegraphics[width=0.50\hsize]{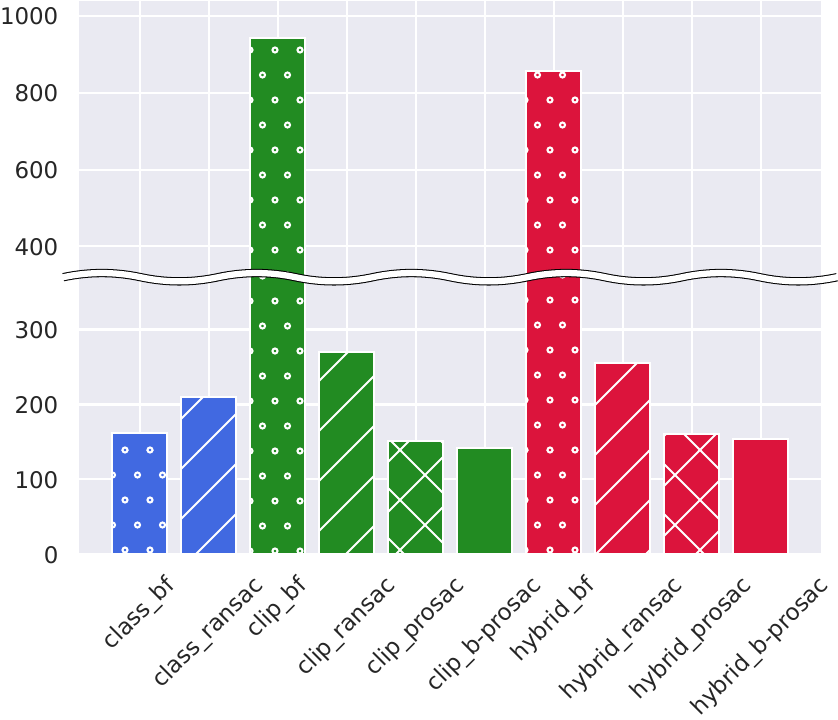}}
	\subfloat[\textit{fr3} \label{fig:runtime_analysis_fr3}]{
		\includegraphics[width=0.50\hsize]{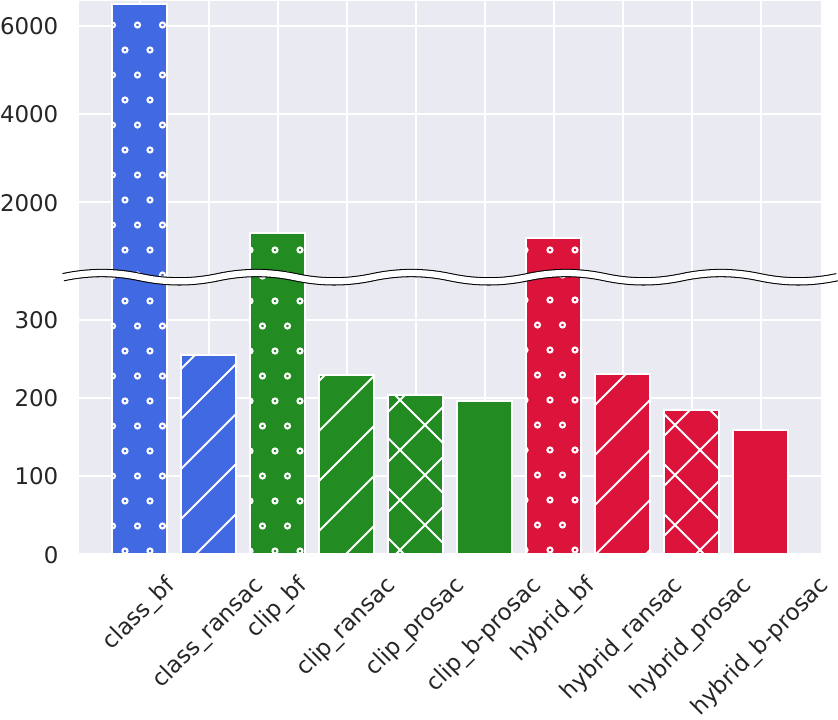}}
	\caption{The average number of iterations required to achieve the final solution
		when the maximum number of iteration is 500. Using the same matching type,
		PROSAC or B-PROSAC consistently found the best solution in earlier iterations.}
	\label{fig:runtime_analysis}
\end{figure}

Fig. \ref{fig:runtime_analysis} shows
the average number of iterations
required to find the best solution in each method.
%We can see that the proposed combination of CLIP-based
%correspondence candidate generation
%and B-PROSAC on average fell into the best solution
%much earlier than the baselines.
\textit{hybrid\_b-prosac} on average fell into the best solution
much earlier than the baselines.
B-PROSAC allows faster convergence
thanks to the sampling strategy considering
the similarity score between an observation and a text label
on the landmark.
This result shows the advantage of the proposed method
over the conventional method
based on object classes and RANSAC.

\subsection{Discussions}

While we confirmed the effectiveness of the proposed method,
there are some factors to be further examined.

\noindent
\textbf{Limitation in the accuracy of CLIP}
%CLIPの推定精度は画像サイズ等に依存し，観測によっては全く正しい対応を求めることができなかった．
%異なる解像度の画像に対しより精度の高いモデルが理想的．
%実際には精度に限りがあるので，使用する観測の取捨選択が必要．
The accuracy of CLIP depends on image quality,
and in some cases, it completely fails to establish correct correspondences.
It is ideal to train a more accurate model. %VLM that can
%correctly recognize images of different resolutions.
In practice, however, we should consider
careful selection of observations for correspondence matching
to increase the robustness of the method
as errors in estimation of machine learning models
is inevitable.

\noindent
\textbf{Dealing with larger scale maps}
%実験が机規模のサイズに留まっているため，より大きな環境での実験が課題．
%現在の枠組みでは全く同じ物体や見えが似通った物体が$k$個以上ある場合に
%対応できない．
%地図上の物体の分布に応じて適応的に$k$の値を変化させるなど，
%より柔軟性の高い手法が必要．
%conducting experiments in larger environments poses a challenge.
The current framework cannot handle cases where there more than $k$ objects that are
either identical or closely resembling each other.
To address this issue, a more flexible approach is required,
such as adapting the value of $k$ based on the landmark distribution.
%As the scale of our experiments is limited to tabletop size,
%examining and improving the scalability of the method are
%important next steps.
%for better scalability of the method.

\noindent
\textbf{Optimality of parameters}
In the experiments, we empirically set the parameter $k=3$
as it yielded the best results overall.
However, how to set the optimal parameter is not trivial
and not investigated well.
An adaptive way of parameter setting is required.

\begin{comment}
\end{comment}

% % 
% % We report
% % the number of iterations necessary to
% % achieve a certain success rate in
% % the ordinary RANSAC-based method
% % and our CLIP-Loc.
% 
% \begin{table}[tb]
% 	\centering
% 	\caption{Parameter analysis}
% 	\label{table:experiment_runtime_analysis}
% 
% 	\begin{tabular}{ c c c }
% 		\toprule
% 		  & RANSAC & \textbf{Proposed} \\
% 		\midrule
% 		N & 10000  & 100               \\
% 		k & -      &                   \\
% 		\bottomrule
% 	\end{tabular}
% \end{table}

%% file: section/5_conclusion.tex
\section{Conclusion and Future Work}
%\section{Conclusion}

We presented a method of landmark association
for global localization in an object map.
It leverages landmarks with
fine-grained information about the landmarks
given as natural language labels,
and the multi-modal capability of CLIP.
By effectively associating the text-labeled landmarks
with visual observations,
the proposed method improves the performance
of global localization.
In addition, we introduced a sampling strategy
inspired by PROSAC
for boosting the efficiency of sampling-based
inlier extraction.
As a result,
we confirmed that the proposed method enables
more accurate and more efficient
camera global localization
compared to the baseline based on
object classes and RANSAC.

% Currently, text labels for the map landmarks
% are given manually to verify
% the feasibility of the proposed algorithm.
As future work,
%we are looking to develop a method
%to automatically build an object map
%with natural language labels.
%Another direction is
we are looking to apply the proposed method to
loop closing capability of
an object SLAM system,
%with a full capability of loop closing,
which has not actively explored yet.
The data association procedure in CLIP-Loc
could be applied to detection of previously
observed landmarks.
\begin{comment}
\end{comment}